\documentclass[runningheads]{llncs}
\usepackage{times}
\usepackage{soul}
\usepackage{url}
\usepackage[hidelinks]{hyperref}
\usepackage[utf8]{inputenc}
\usepackage[small]{caption}
\usepackage{graphicx}
\usepackage{amsmath}
\usepackage{booktabs}
\usepackage{enumitem}
\usepackage{graphicx}
\usepackage{comment}
\usepackage{amssymb}
\usepackage[ruled,vlined,linesnumbered]{algorithm2e}
\begin{document}

\title{On the Tour Towards DPLL(MAPF) and Beyond}

\author{Pavel Surynek\inst{1}\orcidID{0000-0001-7200-0542}}
%

%
\institute{
Faculty of Information Technology\\
Czech Technical University in Prague\\
Th\'{a}kurova 9, 160 00 Praha 6, Czech Republic\\
\email{pavel.surynek@fit.cvut.cz}
}
\maketitle              


\begin{abstract}
We discuss milestones on the tour towards DPLL(MAPF), a {\em multi-agent path finding} (MAPF) solver fully integrated with the Davis–Putnam-Lo\-gemann–Loveland (DPLL) propositional satisfiability testing algorithm through {\em satisfiability modulo theories} (SMT). The task in MAPF is to navigate agents in an undirected graph in a non-colliding way so that each agent eventually reaches its unique goal vertex. At most one agent can reside in a vertex at a time. Agents can move instantaneously by traversing edges provided the movement does not result in a collision. Recently attempts to solve MAPF optimally w.r.t. the sum-of-costs or the makespan based on the reduction of MAPF to propositional satisfiability (SAT) have appeared. The most successful methods rely on building the propositional encoding for the given MAPF instance lazily by a process inspired in the SMT paradigm. The integration of satisfiability testing by the SAT solver and the high-level construction of the encoding is however relatively loose in existing methods. Therefore the ultimate goal of research in this direction is to build the DPLL(MAPF) algorithm, a MAPF solver where the construction of the encoding is fully integrated with the underlying SAT solver. We discuss the current state-of-the-art in MAPF solving and what steps need to be done to get DPLL(MAPF). The advantages of DPLL(MAPF) in terms of its potential to be alternatively parametrized with MAPF$^R$, a theory of continuous MAPF with geometric agents, are also discussed.

\end{abstract}

\section{Introduction}

In {\em multi-agent path finding} (MAPF) \cite{DBLP:conf/focs/KornhauserMS84,DBLP:journals/jair/Ryan08,SharonSFS15,DBLP:journals/ai/SharonSGF13,DBLP:conf/aiide/Silver05,DBLP:conf/icra/Surynek09,DBLP:journals/jair/WangB11} the task is to navigate agents from given starting positions to given individual goals. The standard version of the problem takes place in undirected graph $G=(V,E)$ where agents from set $A=\{a_1,a_2,...,a_k\}$ are placed in vertices with at most one agent per vertex. The initial configuration of agents in vertices of the graph can be written as $\alpha_0: A \rightarrow V$ and similarly the goal configuration as $\alpha_+: A \rightarrow V$. The task of navigating agents can be expressed as a task of transforming the initial configuration of agents $\alpha_0: A \rightarrow V$ into the goal configuration $\alpha_+: A \rightarrow V$.

Movements of agents are instantaneous and are possible across edges into neighbor vertices assuming no other agent is entering the same target vertex. This formulation permits agents to enter vertices being simultaneously vacated by other agents. Trivial case when a pair of agents swaps their positions across an edge is forbidden in the standard formulation. We note that different versions of MAPF exist where for example agents always move into vacant vertices \cite{DBLP:journals/amai/Surynek17}. We usually denote the configuration of agents at discrete time step $t$ as $\alpha_t: A \rightarrow V$. Non-conflicting movements transform configuration $\alpha_t$ {\em instantaneously} into next configuration  $\alpha_{t+1}$. We do not consider what happens between $t$ and $t+1$ in this discrete abstraction. Multiple agents can move at a time hence the MAPF problem is inherently parallel.

In order to reflect various aspects of real-life applications variants of MAPF have been introduced such as those considering {\em kinematic constraints} \cite{DBLP:conf/ijcai/HonigK00XAK17}, {\em large agents} \cite{LargeAAAI2019}, or {\em deadlines} \cite{DBLP:conf/ijcai/0001WFLKK18} - see \cite{DBLP:journals/corr/0001KA0HKUXTS17} for more variants.

\subsection{Lazy Construction of SAT Encodings}

This paper summarizes the development SMT-CBS \cite{Surynek_IJCAI-2019}, a novel optimal MAPF algorithm that {\bf unifies} two major approaches to solving MAPF optimally: a {\bf search-based} approach represented by {\em conflict-based search} (CBS) \cite{SharonSFS15} and a {\bf compilation-based} approach represented by reducing MAPF to propositional satisfiability (SAT) \cite{Biere:2009:HSV:1550723} in the MDD-SAT algorithm \cite{SurynekFSB16}. The SMT-CBS algorithm rephrases ideas of CBS in the terms of {\em satisfiability modulo theories} (SMT) \cite{DBLP:journals/constraints/BofillPSV12} at the high-level. While at the low-level it uses the SAT encoding from MDD-SAT.

Unlike the original CBS that resolves conflicts between agents by branching the search, SMT-CBS refines the propositional model with a disjunctive constraint to resolve the conflict. SMT-CBS hence does not branch at the high-level but instead incrementally extends the propositional model that is consulted with the external SAT solver similarly as it has been done in MDD-SAT.  In contrast to MDD-SAT where the propositional model is fully constructed in a single-shot, the propositional model is being built lazily in SMT-CBS as new conflicts appear.


The hypothesis behind the design of SMT-CBS is that in many cases we do not need to add all constraints to form the {\em complete propositional model} while still be able to obtain a conflict-free solution. Intuitively we expect that such cases where the {\em incomplete propositional model} will suffice are represented by sparsely occupied instances with large environments. The expected benefit in contrast to MDD-SAT is that incomplete model can be constructed and solved faster. On the other hand we expect that the superior performance of MDD-SAT in environments densely populated with agents will be preserved as SMT-CBS will quickly converge the model towards the complete one.

\subsection{Towards DPLL(MAPF) / CDCL(MAPF)}

SMT-CBS represents a milestone towards an optimal MAPF solver where the SAT solver and the high-level construction of the propositional model are fully integrated, an algorithm we denote DPLL(MAPF). DPLL(T)  \cite{DBLP:journals/jacm/NieuwenhuisOT06} commonly denotes an algorithm integrating the SAT solver \cite{DBLP:conf/sat/AudemardLS13} with a decision procedure for the conjunctive fragment of some first-order theory $T$. Similarly as in SMT-CBS the SAT solver decides what literals in a given formula should be set to $\mathit{TRUE}$ in order to satisfy the formula. Subsequently, the decision procedure for the conjunctive fragment checks if the suggested truth value assignment is consistent with $T$. In our case, the theory is represented by movement rules of MAPF and the formula encodes a question if there is a solution of given MAPF of specified value of the objective.

DPLL stands here for the standard search-based SAT solving procedure (Davis\-Putnam-Logemann–Loveland) \cite{DBLP:journals/cacm/DavisLL62} but in fact we use its more modern variants known as CDCL (Conflict-Driven Clause Learning) \cite{DBLP:journals/tc/Marques-SilvaS99}, hence precisely our algorithm should be denoted CDCL(MAPF) however we will keep the more common notation DPLL(MAPF) used in the literature.

The {\bf organization} of the paper is as follows: We first introduce MAPF formally. Then the combination of CBS and MDD-SAT, the recent optimal MAPF algorithm SMT-CBS is recalled. On top of this we discuss the future perspective of DPLL(MAPF) in more details.

\section{Multi-Agent Path Finding Formally}

The {\em Multi-agent path finding} (MAPF) problem \cite{DBLP:conf/aiide/Silver05,DBLP:journals/jair/Ryan08} consists of an undirected graph $G=(V,E)$ and a set of agents $A=\{a_1, a_2, ..., a_k\}$ such that $|A| \leq |V|$. Each agent is placed in a vertex so that at most one agent resides in each vertex. The placement of agents is denoted $\alpha: A \rightarrow V$. Next we are given initial configuration of agents $\alpha_0$ and goal configuration $\alpha_+$.

\begin{figure}[h]
    \centering
    \includegraphics[trim={3.0cm 24cm 3.5cm 2.8cm},clip,width=0.9\textwidth]{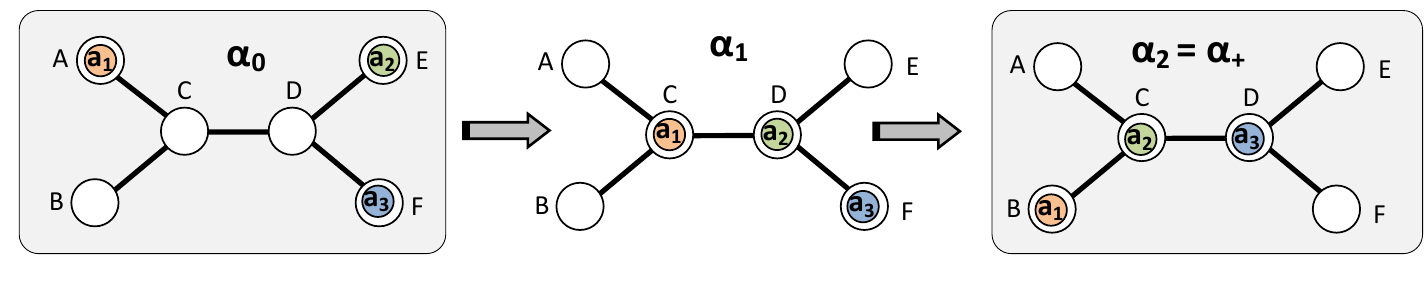}
    \vspace{-0.2cm}\caption{A MAPF instance with three agents $a_1$, $a_2$, and $a_3$. A two-step solution is shown too.}
    \label{figure-MAPF}
\end{figure}

At each time step an agent can either {\em move} to an adjacent vertex or {\em wait} in its current vertex. The task is to find a sequence of move/wait actions for each agent $a_i$, moving it from $\alpha_0(a_i)$ to $\alpha_+(a_i)$ such that agents do not {\em conflict}, i.e., do not occupy the same vertex at the same time nor cross the same edge in opposite directions simultaneously. The following definition formalizes the commonly used movement rule in MAPF. An example of MAPF instance is shown in Figure \ref{figure-MAPF}.

\begin{definition}
    {\bf Valid movement in MAPF.}
    Configuration $\alpha'$ results from $\alpha$ if and only if the following conditions hold:
    
    \begin{enumerate}[label=(\roman*)]
      \item {$\alpha(a) = \alpha'(a)$ or $\{\alpha(a),\alpha'(a)\} \in E$ for all $a \in A$ (agents wait or move along edges);}
      \item {for all $a \in A$ it holds ${\alpha(a) \neq \alpha'(a)} \Rightarrow {\neg (\exists b \in A)(\alpha(b) = \alpha'(a) \wedge \alpha'(b) = \alpha(a))}$ (no two agents crosses an edge in opposite directions);}
      \item {and for all $a,a' \in A$ it holds that ${a \neq a'} \Rightarrow {\alpha'(a) \neq \alpha'(a')}$ (no two agents share a vertex in the next configuration)}.
    \end{enumerate}
    \label{def:movement}
    \vspace{-0.2cm}
\end{definition}

Solving MAPF is to find a sequence of configurations $[\alpha_0,\alpha_1,...,\alpha_{\mu}]$ such that  $\alpha_{i+1}$ results using valid movements from $\alpha_{i}$ for $i=1,2,...,\mu-1$, and $\alpha_{\mu}=\alpha_+$. A {\em feasible solution} of a solvable MAPF instance can be found in polynomial time \cite{WILSON197486,DBLP:conf/focs/KornhauserMS84}; precisely the worst case time complexity of most practical algorithms for finding feasible solutions is $\mathcal{O}({|V|}^3)$ \cite{LunaB11,DBLP:journals/jair/WildeMW14}.

We are often interested in optimal solutions. In case of the {\em makespan} \cite{DBLP:journals/amai/Surynek17} we just need to minimize $\mu$ in the aforementioned solution sequence. For introducing the {\em sum-of-costs} objective \cite{dresner2008aMultiagent,standley2010finding,DBLP:journals/ai/SharonSGF13} we need more notation as follows:

\begin{definition}
	{\bf Sum-of-costs objective} is the summation, over all $k$ agents, of the number of time steps required to reach the goal vertex.
	Denoted $\xi$, where $\xi = \sum_{i=1}^k{\xi(\mathit{path}(a_i))}$ and $\xi( \mathit{path}(a_i))$ is an \textit{individual path cost} of agent $a_i$
	connecting $\alpha_0(a_i)$ and $\alpha_+(a_i)$  calculated as the number of edge traversals and wait actions. \footnote{The notation $ \mathit{path}(a_i)$ refers to
	path in the form of a sequence of vertices and edges connecting $\alpha_0(a_i)$ and $\alpha_+(a_i)$ while $\xi$ assigns the cost to a
	given path.}
\end{definition}



We note that finding a solution that is optimal (minimal) with respect to either the makespan or the sum-of-costs objective is NP-hard \cite{DBLP:conf/aaai/RatnerW86,DBLP:conf/aaai/Surynek10}. 


\section{Unifying Search-based and Compilation-based Approaches}

Before introducing SMT-CBS, a unification between the search-based and the compilation-based approach we will briefly discuss both approaches themselves.

\subsection{Conflict-based Search}

CBS is a representative of {\bf search-based approach}. CBS uses the idea of resolving conflicts lazily; that is, a solution of MAPF instance is not searched against the complete set of movement constraints that forbids collisions between agents but with respect to initially empty set of collision forbidding constraints that gradually grows as new conflicts appear. The advantage of CBS is that it can find a valid solution before all constraints are added.

The high-level of CBS searches a {\em constraint tree} (CT) using a priority queue in breadth first manner. CT is a binary tree where each node $N$ contains a set of collision avoidance constraints $N.\mathit{constraints}$ - a set of triples $(a_i,v,t)$ forbidding occurrence of agent $a_i$ in vertex $v$ at time step $t$, a solution $N.paths$ - a set of $k$ paths for individual agents, and the total cost $N.\xi$ of the current solution.

The low-level process in CBS associated with node $N$ searches paths for individual agents with respect to set of constraints $N.\mathit{constraints}$. For a given agent $a_i$, this is a standard single source shortest path search from $\alpha_0(a_i)$ to $\alpha_+(a_i)$ that avoids a set of vertices $\{v \in V|(a_i,v,t) \in N. \mathit{constraints}\}$ whenever working at time step $t$. For details see \cite{SharonSFS15}.

CBS stores nodes of CT into priority queue $\textsc{Open}$ sorted according to the ascending costs of solutions. At each step CBS takes node $N$ with the lowest cost from $\textsc{Open}$ and checks if $N.\mathit{paths}$ represent paths that are valid with respect to MAPF movements rules - that is, $N.\mathit{paths}$ are checked for collisions. If there is no collision, the algorithms returns valid MAPF solution $N.\mathit{paths}$. Otherwise the search branches by creating a new pair of nodes in CT - successors of $N$. Assume that a collision occurred between agents $a_i$ and $a_j$ in vertex $v$ at time step $t$. This collision can be avoided if either agent $a_i$ or agent $a_j$ does not reside in $v$ at timestep $t$. These two options correspond to new successor nodes of $N$ - $N_1$ and $N_2$ that inherit the set of conflicts from $N$ as follows: $N_1.\mathit{conflicts} = N.\mathit{conflicts} \cup \{(a_i,v,t)\}$ and $N_2.\mathit{conflicts} = N.\mathit{conflicts} \cup \{(a_j,v,t)\}$. $N_1.\mathit{paths}$ and $N_1.\mathit{paths}$ inherit paths from $N.\mathit{paths}$ except those for agents $a_i$ and $a_j$ respectively. Paths for $a_i$ and $a_j$ are recalculated with respect to extended sets of conflicts $N_1.\mathit{conflicts}$ and $N_2.\mathit{conflicts}$ respectively and new costs for both agents $N_1.\xi$ and $N_2.\xi$ are determined. After this, $N_1$ and $N_2$ are inserted into the priority queue $\textsc{Open}$.

The pseudo-code of CBS is listed as Algorithm \ref{alg-CBS}. One of crucial steps occurs at line 16 where a new path for colliding agents $a_i$ and $a_j$ is constructed with respect to the extended set of conflicts. $N.paths(a)$ refers to path of agent $a$.

\begin{algorithm}[t!]
\begin{footnotesize}
\SetKwBlock{NRICL}{CBS ($G=(V,E),A,\alpha_0,\alpha_+)$}{end} \NRICL{
    $R.\mathit{constraints} \gets \emptyset$ \\
    $R.\mathit{paths} \gets$ $\{$shortest path from $\alpha_0(a_i)$ to $\alpha_+(a_i) | i = 1,2,...,k\}$\\
    $R.\xi \gets \sum_{i=1}^k{\xi(N.\mathit{paths}(a_i))}$ \\
    insert $R$ into $\textsc{Open}$ \\
    \While {$\textsc{Open} \neq \emptyset$} {
        $N \gets$ min($\textsc{Open}$)\\
        remove-Min($\textsc{Open}$)\\
        $\mathit{collisions} \gets$ validate($N.paths$)\\
        \If {$\mathit{collisions} = \emptyset$}{
            \Return $N.\mathit{paths}$\\
        }
        let $(a_i,a_j,v,t) \in \mathit{collisions}$\\
        
        \For {each $a \in \{a_i,a_j\}$}{
       	$N'.\mathit{constraints} \gets N.\mathit{constraints} \cup \{(a,v,t)\}$\\
        	$N'.\mathit{paths} \gets N.\mathit{paths}$\\
        	update($a$, $N'.\mathit{paths}$, $N'.\mathit{constraints}$)\\
		$N'.\xi \gets \sum_{i=1}^k\xi{(N'.\mathit{paths}(a_i))}$\\
		insert $N'$ into $\textsc{Open}$ \\
        }
     }
} \caption{CBS algorithm for MAPF solving} \label{alg-CBS}
\end{footnotesize}
\end{algorithm}

\subsection{Compilation to Propositional Satisfiability}

The major alternative to CBS is represented by {\bf compilation} of MAPF to propositional satisfiability (SAT) \cite{SurynekFSB16,DBLP:journals/amai/Surynek17}. The idea follows SAT-based planning \cite{DBLP:conf/ijcai/KautzS99} where the existence of a plan for a fixed number time steps is modeled as SAT. We similarly construct propositional formula $\mathcal{F(\xi)}$ such that it is satisfiable if and only if a solution of a given MAPF of sum-of-costs $\xi$ exists. Moreover, the approach is constructive; that is, $\mathcal{F(\xi)}$ exactly reflects the MAPF instance and if satisfiable, solution of MAPF can be reconstructed from satisfying assignment of the formula. We say  $\mathcal{F(\xi)}$ to be a {\em complete propositional model} of MAPF.

\begin{definition}
  {\bf (complete propositional model).} Propositional formula $\mathcal{F(\xi)}$ is a {\em complete propositional model} of MAPF $\Sigma$
  if the following condition holds:
  \begin{center}
  $\mathcal{F(\xi)}$ is satisfiable $\Leftrightarrow$ $\Sigma$ has a solution of sum-of-costs $\xi$.
  \end{center}
\end{definition}

Being able to construct such formula $\mathcal{F}$ one can obtain optimal MAPF solution by checking satisfiability of $\mathcal{F}(\xi_0)$, $\mathcal{F}(\xi_0+1)$, $\mathcal{F}(\xi_0+2)$,... until the first satisfiable $\mathcal{F(\xi)}$ is met ($\xi_0$ is the lower bound for the sum-of-costs calculated as the sum of lengths of shortest paths). This is possible due to monotonicity of MAPF solvability with respect to increasing values of common cumulative objectives. Details of $\mathcal{F}$ are given in \cite{SurynekFSB16}.

The advantage of the SAT-based approach is that state-of-the-art SAT solvers can be used for determining satisfiability of $\mathcal{F}(\xi)$ \cite{DBLP:conf/ijcai/AudemardS09}.

\section{Combining SMT and CBS}

A natural relaxation from the complete propositional model is an {\em incomplete propositional model} where instead of the equivalence between solving MAPF and the formula we require an implication only.

\begin{definition}
  {\bf (incomplete propositional model).} Propositional formula $\mathcal{H(\xi)}$ is an {\em incomplete propositional model} of MAPF 
  $\Sigma$ if the following condition holds:
  \begin{center}
  $\mathcal{H(\xi)}$ is satisfiable $\Leftarrow$ $\Sigma$ has a solution of sum-of-costs $\xi$.
  \end{center}
\end{definition}

A close look at CBS reveals that it operates similarly as problem solving in {\em satisfiability modulo theories} (SMT) \cite{DBLP:journals/constraints/BofillPSV12}. SMT divides satisfiability problem in some complex theory $T$ into an abstract propositional part that keeps the Boolean structure of the decision problem and a simplified decision procedure $\mathit{DECIDE_T}$ that decides fragment of $T$ restricted on {\em conjunctive formulae}. A general $T$-formula $\Gamma$ is transformed to a {\em propositional skeleton} by replacing atoms with propositional variables. The SAT solver then decides what variables should be assigned $\mathit{TRUE}$ in order to satisfy the skeleton - these variables tells what atoms hold in $\Gamma$. $\mathit{DECIDE_T}$ then checks if the conjunction of atoms assigned $\mathit{TRUE}$ is valid with respect to axioms of $T$. If so then satisfying assignment is returned. Otherwise a conflict from $\mathit{DECIDE_T}$ (often called a lemma) is reported back and the skeleton is extended with a constraint forbidding the conflict.

The above observation led us to the idea to rephrase CBS in terms of SMT. The abstract propositional part working with the skeleton will be taken from MDD-SAT provided that only constraints ensuring that assignments form valid paths interconnecting starting positions with goals will be preserved. Other constraints for collision avoidance will be omitted initially. This will result in an {\em incomplete propositional model}.

The paths validation procedure will act as $\mathit{DECIDE_T}$ and will report back the set of conflicts found in the current solution. Hence axioms of $T$ will be represented by the movement rules of MAPF. We call the resulting algorithm SMT-CBS and it is shown in pseudo-code as Algorithm \ref{alg-SMTCBS}.

\begin{algorithm}[h]
\begin{footnotesize}
\SetKwBlock{NRICL}{SMT-CBS ($\Sigma = (G=(V,E),A,\alpha_0,\alpha_+))$}{end} \NRICL{
    $\mathit{conflicts} \gets \emptyset$\\
    $\mathit{paths} \gets$ $\{\mathit{path}^*(a_i)$ a shortest path from $\alpha_0(a_i)$ to $\alpha_+(a_i) | i = 1,2,...,k\}$ \\
    $\xi \gets \sum_{i=1}^k{\xi(\mathit{paths}(a_i))}$ \\
    \While {$\mathit{TRUE}$}{
         $(\mathit{paths,conflicts}) \gets$ SMT-CBS-Fixed($\mathit{conflicts},\xi,\Sigma$)\\
        \If {$\mathit{paths} \neq$ UNSAT}{
        	\Return $\mathit{paths}$\\
        }
        $\xi \gets \xi + 1$\\
    }
}   
 
\SetKwBlock{NRICL}{SMT-CBS-Fixed($conflicts,\xi,\Sigma$)}{end} \NRICL{
	    $\mathcal{H}(\xi) \gets$ encode-Basic$(\mathit{conflicts},\xi,\Sigma)$\\
	    \While {$\mathit{TRUE}$}{
	        $\mathit{assignment} \gets$ consult-SAT-Solver$(\mathcal{H}(\xi))$\\
	        \If {$\mathit{assignment} \neq \mathit{UNSAT}$}{
	            $\mathit{paths} \gets$ extract-Solution$(\mathit{assignment})$\\
	            $\mathit{collisions} \gets$ validate($\mathit{paths}$)\\
                   \If {$\mathit{collisions} = \emptyset$}{
                      \Return $(\mathit{paths,conflicts})$\\
                   }
                   \For{each $(a_i,a_j,v,t) \in \mathit{collisions}$}{
                      $\mathcal{H}(\xi) \gets \mathcal{H}(\xi) \cup \{\neg \mathcal{X}_v^t(a_i) \vee \neg \mathcal{X}_v^t(a_j)$\}\\
                      $\mathit{conflicts} \gets \mathit{conflicts} \cup \{[(a_i,v,t),(a_j,v,t)]\}$
                   }
               }
               \Return {(UNSAT,$\mathit{conflicts}$)}\\
          }
}
\caption{SMT-CBS algorithm for MAPF solving} \label{alg-SMTCBS}
\end{footnotesize}
\end{algorithm}

The algorithm is divided into two procedures: SMT-CBS representing the main loop and SMT-CBS-Fixed solving the input MAPF for fixed cost $\xi$. The major difference from the standard CBS is that there is no branching at the high-level. The high-level SMT-CBS roughly correspond to the main loop of MDD-SAT. The set of conflicts is iteratively collected during the entire execution of the algorithm. Procedure {\em encode} from MDD-SAT is replaced with {\em encode-Basic} that produces encoding that ignores specific movement rules (collisions between agents) but in contrast to {\em encode} it encodes collected conflicts into $\mathcal{H}(\xi)$.

The conflict resolution in the standard CBS implemented as high-level branching is here represented by refinement of $\mathcal{H}(\xi)$ with disjunction (line 20). The presented SMT-CBS can eventually build the same formula as MDD-SAT but this is done lazily in SMT-CBS.

\section{From SMT-CBS to DPLL(MAPF) and Beyond}

Although the performed experimental evaluation presented in \cite{Surynek_IJCAI-2019} shows that SMT-CBS significantly outperforms both the CBS algorithm and MDD-SAT we conjecture that there is still room for improving the idea behind SMT-CBS. SMT-CBS as shown in Algorithm \ref{alg-SMTCBS} implements very basic variant of SMT-based problem solving. More advanced SMT-based algorithms for deciding formulae in first order theories integrates the SAT solver and $\mathit{DECIDE_T}$ more tightly. $\mathit{DECIDE_T}$ is invoked not only for the fully assigned formulae (line 13 produces a full assignment) but also for partial assignments. Such algorithms are usually denoted DPLL(T) \cite{DBLP:journals/jacm/NieuwenhuisOT06}.

Whenever the SAT-solving search loop in DPLL (or CDCL) assigns new propositional variable, $\mathit{DECIDE_T}$ is invoked to check if the extended partial assignment is consistent with respect to $T$ (or MAPF in our case). If so the main loop continues with assigning the next variable. Otherwise $\mathit{DECIDE_T}$ return a lemma forbidding the current assignment, translated in MAPF terms this corresponds to a case when $\mathit{DECIDE_{MAPF}}$ discovers a conflict and returns the conflict elimination constraint.

$\mathit{DECIDE_T}$ ($\mathit{DECIDE_{MAPF}}$) can do much more in the consistent case. The procedure can derive new assignments that is to perform a form of MAPF constraint propagation. Such propagation could work in tandem with the standard {\em unit propagation} \cite{DBLP:journals/jlp/DowlingG84} integrated in the CDCL-based SAT solvers. Moreover DPLL(MAPF) algorithm can be further parametrized with variable and value selection heuristics that can take into account axioms of the MAPF theory.  

\subsection{Implications for MAPF with Continuous Time and Geometric Agents}

Recently generalizations of MAPF considering continuous time and geometric agents have appeared \cite{DBLP:journals/corr/abs-1901-05506}. The continuous variant denoted MAPF$^\mathcal{R}$ assigns each vertex of the underlying graph from the standard MAPF coordinates in the Euclidean space. Agents are geometric objects such as circles, spheres, polygons etc. and can move along straight lines connecting vertices of the underlying graph (agents cannot move outside these predefined lines). For simplicity agents are assumed to have constant speeds but the algorithmic concepts generalize for non-constant models too.

Collisions between agents are defined as any overlap between their bodies. The task is to find a collision free temporal plan. An adaptation of CBS algorithm denoted CBS$^\mathcal{R}$ for the continuous variant has been proposed \cite{DBLP:journals/corr/abs-1901-05506}. We rephrased CBS$^\mathcal{R}$ in terms of SMT similarly as it has been done with CBS. The resulting algorithm SMT-CBS$^\mathcal{R}$ \cite{DBLP:journals/corr/abs-1903-09820} differs from SMT-CBS mostly in the aspect of decision variable generation. In case of the standard MAPF, decision variables can be determined statically in advance but this is not applicable in the continuous version since the continuity has potential to spawn infinitely many decisions. Decision variables need to be generated dynamically in response to discoveries of new conflicts in MAPF$^\mathcal{R}$.

Naturally the future step in the development of SMT-CBS$^\mathcal{R}$ is DPLL(MAPF$^\mathcal{R}$) where also partial assignments will be checked for consistency.


\section{Discussion and Conclusion}

We recalled the new MAPF solving method called SMT-CBS that combines search-based CBS and SAT-based MDD-SAT through concepts from satisfiability modulo theories (SMT). Although SMT-CBS represent state-of-the-art in optimal MAPF solving as shown in \cite{Surynek_IJCAI-2019} we identified certain deficiencies of the algorithm. Namely the fact that it checks for consistency with respect to MAPF rules only fully assigned propositional encodings. We therefore theoretically suggest DPLL(MAPF), a new algorithm that will also check for consistency partial assignments. We further suggest future work in which DPLL(MAPF) will be generalized for MAPF with continuous time and geometric agents, an analogous next step beyond the existing SMT-CBS$^\mathcal{R}$ algorithm \cite{DBLP:journals/corr/abs-1903-09820}.

\section*{Acknowledgements}
\noindent 
This research has been supported by GA\v{C}R - the Czech Science Foundation, grant registration number 19-17966S.

\bibliographystyle{splncs04}
\bibliography{bibfile}

\end{document}